\documentclass[conference]{IEEEtran}
\IEEEoverridecommandlockouts
\usepackage{cite}
\usepackage{algorithm,algorithmicx,algpseudocode}
\usepackage{amsmath,amssymb,amsfonts,amsthm}
\usepackage{graphicx}
\usepackage{textcomp}
\usepackage{mathrsfs}
\usepackage[mathscr]{eucal}
\def\BibTeX{{\rm B\kern-.05em{\sc i\kern-.025em b}\kern-.08em
    T\kern-.1667em\lower.7ex\hbox{E}\kern-.125emX}}
\usepackage{hyperref}
\usepackage{stfloats}
\usepackage{booktabs}
\usepackage{ulem}
\usepackage{multirow, multicol}
\usepackage{xcolor}
\usepackage{adjustbox}
\usepackage{caption}
\usepackage{hyperref}

\DeclareRobustCommand*{\IEEEauthorrefmark}[1]{%
    \raisebox{0pt}[0pt][0pt]{\textsuperscript{\footnotesize\ensuremath{#1}}}}

\begin{document}
\normalem

\title{Robust Semi-Supervised Learning for Self-learning Open-World Classes}

\author{
    \IEEEauthorblockN{
        Wenjuan Xi\IEEEauthorrefmark{1},
        Xin Song\IEEEauthorrefmark{2},
        Weili Guo\IEEEauthorrefmark{1}$^{,\dag}$, 
        Yang Yang\IEEEauthorrefmark{1,3}$^{,\dag}$ \thanks{$^{\dag}$ Corresponding authors.}}
    \IEEEauthorblockA{\IEEEauthorrefmark{1}Nanjing University of Science and Technology, China}
    \IEEEauthorblockA{\IEEEauthorrefmark{2}Baidu Talent Intelligence Center, China}
    \IEEEauthorblockA{\IEEEauthorrefmark{3}Hong Kong Polytechnic University, China}
    {\{xiwenjuan, wlguo, yyang\}}@njust.edu.cn, songxin06@baidu.com
}


\maketitle

\begin{abstract}
    Existing semi-supervised learning (SSL) methods assume that labeled and unlabeled data share the same class space. However, in real-world applications, unlabeled data always contain classes not present in the labeled set, which may cause classification performance degradation of known classes. Therefore, open-world SSL approaches are researched to handle the presence of multiple unknown classes in the unlabeled data, which aims to accurately classify known classes while fine-grained distinguishing different unknown classes. To address this challenge, in this paper, we propose an open-world SSL method for Self-learning Open-world Classes (SSOC), which can explicitly self-learn multiple unknown classes. Specifically, SSOC first defines class center tokens for both known and unknown classes and autonomously learns token representations according to all samples with the cross-attention mechanism. To effectively discover novel classes, SSOC further designs a pairwise similarity loss in addition to the entropy loss, which can wisely exploit the information available in unlabeled data from instances' predictions and relationships. Extensive experiments demonstrate that SSOC outperforms the state-of-the-art baselines on multiple popular classification benchmarks. Specifically, on the ImageNet-100 dataset with a novel ratio of 90\%, SSOC achieves a remarkable 22\% improvement.
\end{abstract}


\section{Introduction}

    \begin{figure}[htbp]
      \centering
      \includegraphics[trim=0 0 0 0, clip, width=0.5\textwidth]{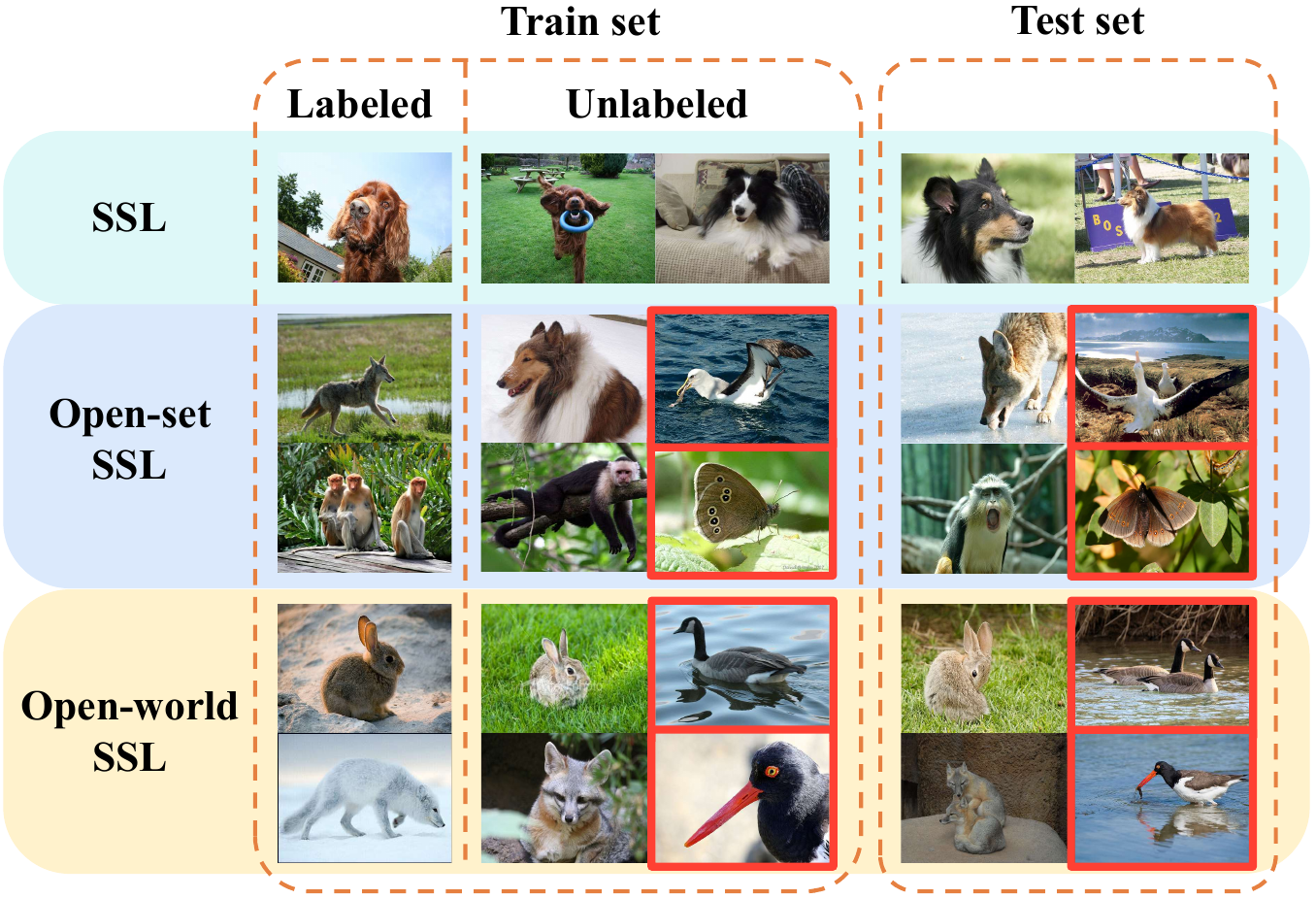}
      \caption{SSL settings do not include the concept of novel classes. In open-set SSL, the unknown classes in the unlabeled dataset are treated as a single category, and exclusively focus on known class classification. In open-world SSL, multiple unknown classes can be distinguished. Unknown class samples are highlighted with a red border.}
      \label{fig:intro}
    \end{figure}
    
    With the development of machine learning, deep learning has achieved significant accomplishments in various domains such as vision, text, and speech \cite{aggarwal2022has}, \cite{DBLP:journals/nn/MatsuoLSPSSUM22}, \cite{DBLP:journals/csur/PouyanfarSYTTRS19}, \cite{dargan2020survey}. In the early stages, supervised learning relied on a large amount of expensive labeled data for model training. SSL emerged to reduce costs by effectively utilizing abundant unlabeled data \cite{DBLP:journals/tkde/YangFZLJ21}, achieving performance comparable to supervised learning in scenarios with limited labeled data \cite{DBLP:journals/tkde/YangZZXJY23}, \cite{DBLP:journals/ml/EngelenH20}. However, almost all SSL methods are based on a default assumption: unlabeled data and labeled data are sampled from the same distribution, and there are no unseen classes for the model, as shown in the first row of Fig. \ref{fig:intro}. While this assumption has enabled their widespread application in closed-world scenarios \cite{DBLP:journals/tkde/YangZWLXJ21}, \cite{DBLP:journals/corr/abs-2110-11767}, it leads to significant performance degradation in real-world situations \cite{DBLP:conf/icml/PezeshkiFBCB16}, \cite{DBLP:conf/nips/SajjadiJT16}, \cite{DBLP:conf/iclr/LaineA17}, \cite{DBLP:conf/nips/TarvainenV17}. For instance, in pathological image analysis, some lesion tissue slide images may come from unknown diseases, each with significant variations. This requires models that can distinguish unknown pathological image categories. In the field of network security, security experts use models for detecting malicious software, with the expectation that these models can differentiate new types of malicious software \cite{ZhouYZ22}, \cite{DBLP:conf/nips/RasmusBHVR15}. Therefore, it is essential to investigate a more inclusive and versatile open-world approach.

    Recently, researchers have proposed the concept of open-world SSL \cite{DBLP:conf/iclr/CaoBL22}. This framework allows for the presence of unknown classes in the unlabeled dataset, which are classes that do not appear in the labeled set, as shown in Fig. \ref{fig:intro}, third row. The objective is to classify both known and unknown classes simultaneously. Previously, open-set SSL methods have been proposed to address the real scenario \cite{geng2020recent}, \cite{DBLP:journals/pami/ScheirerRSB13}. However, unlike the open-world SSL approach discussed in this paper, open-set SSL methods simply reject unknown class samples to prevent them from negatively affecting the classification performance of known classes in the unlabeled dataset. Their primary objective remains classifying the known classes, as depicted in the second row of Figure 1. Similarly, novel class discovery methods \cite{troisemaine2023novel}, \cite{DBLP:conf/ijcnn/NodetLBCO21a}, \cite{zhou2018brief} also consider the open-world setting, but they assume that the unlabeled dataset contains only unknown class samples and focus solely on clustering the unknown classes without considering the model's performance on known classes. Compared to these two approaches, open-world SSL poses a more challenging problem. The key to addressing this problem lies in enabling the model to learn multiple unknown classes better while ensuring the classification performance of known classes.

    The problem of open-world SSL has attracted extensive attention, and several methods have been proposed to address this issue \cite{DBLP:conf/iclr/CaoBL22}, \cite{DBLP:conf/nips/GuoZWSL22}. These methods primarily approach the problem from the perspective of loss functions, employing optimization objectives that facilitate the learning of unknown classes and incorporating uncertainty mechanisms or adaptive thresholds to alleviate the issue of class imbalance during the learning process \cite{DBLP:conf/aaai/ChenZLG20}. Although these methods have achieved remarkable performance in open-world SSL, they are limited to implicitly modeling multiple novel classes and partitioning the novel class space at the label level. This approach makes it difficult for the model to truly understand category concepts, leading to biases and typically exhibiting poor robustness when facing unknown classes. Therefore, there is still a lack of an open-world SSL method that can explicitly model category concepts, effectively discriminate between categories, and uncover hidden patterns and structures in the data.

    To address this, we propose an open-world SSL method called Self-Supervised Open-World Class (SSOC) that aims to explicitly self-learn multiple unknown classes. Specifically, we initialize prototypes (class centers) representations for both known and unknown classes. Then, we iteratively learn the representation of class centers by combining a cross-attention mechanism with data features to achieve explicit modeling of class information. To assist the learning of multiple novel classes, we employ confident unlabeled samples to constrain the entropy loss and utilize pairwise similarity loss to extract information from unlabeled data, ensuring consistency at both the instance-level and prediction-level representations. The model architecture of SSOC not only ensures the classification performance of known classes but also focuses on capturing the differences and similarities among classes. By learning the feature representations of class prototypes, SSOC can perform classification operations with clear physical interpretations. This provides a new direction for exploring the application of interpretable learning methods in open-world SSL tasks.

    In summary, our work has the following contributions:
    \begin{itemize}
        \item We propose an open-world SSL method for self-learning open-world classes. The method exploits a cross-attention mechanism to model class concepts explicitly and learn multiple unknown classes autonomously.
        \item We design a paired similarity loss to intelligently utilize information from unlabeled data for novel class discovery via instance prediction and relationship identification.
        \item We conducted experiments on CIFAR-10, CIFAR-100, and ImageNet-100 datasets with various data partitions to illustrate the effectiveness of SSOC. Our results showcase the remarkable robustness of SSOC in scenarios with limited labeled data and many novel classes.
    \end{itemize}

\section{Related work}
    Open-world SSL is closely related to SSL, open-set SSL, and novel class discovery (NCD). In this section, we summarize the similarities and differences among these research directions and investigate their development history.
    
    \subsection{Semi-Supervised Learning}
        The goal of SSL is to improve the learning performance of models by utilizing a large amount of unlabeled data along with a small amount of labeled data, thereby addressing the problem of high annotation costs. In recent years, significant progress has been made in deep SSL methods, which can be categorized into two main approaches: consistency regularization \cite{DBLP:conf/cvpr/LeeKKCCH22} and pseudo-labeling methods \cite{yang2022survey}, \cite{DBLP:conf/nips/OliverORCG18}. Consistency regularization is based on the core idea that the model's predictions should remain consistent or similar under small perturbations of input data. It aims to enforce the model to produce consistent outputs for perturbed versions of the same input, promoting robustness and generalization. On the other hand, pseudo-labeling methods aim to augment the labeled dataset by using the model's predictions on unlabeled data as pseudo-labels. These pseudo-labels are treated as ground-truth labels for the unlabeled data, expanding the training dataset and enabling the model to learn from the additional unlabeled samples. Many methods combine both ideas. For example, MixMatch \cite{DBLP:conf/nips/BerthelotCGPOR19} applies random data augmentation to unlabeled samples and uses MixUp \cite{DBLP:conf/iclr/ZhangCDL18} pseudo-labeling to leverage information from unlabeled data. FixMatch \cite{DBLP:conf/nips/SohnBCZZRCKL20} uses strong augmented images along with pseudo-labels generated from weakly augmented versions to learn consistency targets. However, SSL methods tend to misclassify unknown samples as known classes, which leads to a significant performance degradation in open-world scenarios. Consequently, traditional SSL methods are not suitable for effectively addressing open-world problems.
        

    \subsection{Open-World Semi-Supervised Learning}
        To address the emergence of novel classes, researchers have started studying the open-world problem. Initially, unknown classes were treated as a single category, leading to the proposal of the open-set SSL scenario. This scenario assumes that the unlabeled dataset may contain samples from unknown classes, and the goal is to reduce the negative impact of unknown classes on the learning of known classes and improve the model's robustness. In recent years, several open-set SSL methods have emerged, focusing on reducing the weight of unknown class samples \cite{guo2020safe} or selectively using unknown class samples \cite{YangWSLZXY22}, \cite{DBLP:conf/cvpr/SunYZLP20}. For example, $\rm DS^3L$ \cite{guo2020safe} decreases the weight of unknown class samples to reduce interference from unknown classes, while S2OSC \cite{YangWSLZXY22} combines the sampling of novel class samples and model retraining in an adaptive manner, effectively incorporating semi-supervised learning. However, in practical applications, there may be multiple novel classes, and research has proposed adding post-processing steps for unknown class clustering to distinguish multiple novel classes after open-set SSL methods \cite{DBLP:conf/iclr/CaoBL22}. However, this approach does not yield satisfactory results due to the lack of focus on unknown classes during training.

        To address these issues, a recent work \cite{DBLP:conf/iclr/CaoBL22} introduces the open-world SSL setting, which assumes the presence of multiple unknown classes in the unlabeled dataset. The goal is to classify known classes while discovering multiple unknown classes. Compared to previous research, the open-world SSL scenario is closer to real-world situations but is still in its early stages. ORCA \cite{DBLP:conf/iclr/CaoBL22} is the first end-to-end deep learning framework proposed to tackle this problem. It establishes an uncertainty-adaptive margin mechanism to enhance the learning of unknown classes and achieves outstanding results. NACH \cite{DBLP:conf/nips/GuoZWSL22} designs adaptive thresholds to balance the learning of known and unknown classes and proposes a novel classification loss function to aid the model in learning unknown classes, achieving better performance than ORCA.

        
    \subsection{Novel Class Discovery}
        The open-world SSL task's essence is discovering multiple unknown novel classes, closely related to NCD, which falls under weakly supervised learning \cite{troisemaine2023novel}, \cite{DBLP:conf/ijcnn/NodetLBCO21a}, \cite{zhou2018brief}, \cite{YangSZFZXY23}. It involves training with both labeled and unlabeled datasets and assumes that both the unlabeled training and test sets comprise samples from unknown classes, with the goal of clustering these unknown classes. Initially, NCD methods typically employ a two-stage learning approach, where prior knowledge is learned on the labeled dataset, and then transfer learning-like methods are used to cluster the unknown classes in the unlabeled set. For instance, CILF \cite{YangSZFZXY23} proposed a framework for incremental learning that acquires adaptive embeddings to tackle novel class detection, while DTC \cite{DBLP:conf/iccv/HanVZ19} proposed a deep transfer clustering method that leverages prior knowledge from known classes to enhance the representation of unknown classes and learns class prototypes to accomplish the clustering of unknown classes. However, NCD methods solely concentrate on the clustering performance of unknown classes while disregarding the classification task of known classes. Consequently, when both known and unknown classes are present in the test set, NCD methods struggle to achieve excellent overall performance.


\section{Preliminary}
    In the context of open-world SSL, the training set consists of a labeled dataset $\mathcal{D}^l={\{(x_i, y_i)\}}_{i=1}^\mathcal{M}$ containing $\mathcal{M}$ labeled samples and an unlabeled dataset $\mathcal{D}^u={\{x_i\}}_{i=1}^\mathcal{N}$ containing $\mathcal{N}$ unlabeled samples, where $x \in \mathbb{R}^D$ and $D$ represents the image input dimension. We define $\mathcal{C}^l$ as the set of classes that appear in $\mathcal{D}^l$, and $\mathcal{C}^u$ as the set of classes that appear in $\mathcal{D}^u$. In the labeled dataset $\mathcal{D}^l$, $y \in \mathcal{C}^l={\{1,...,|\mathcal{C}^l|\}}$, while in the unlabeled dataset $\mathcal{D}^u$, $x$ belongs to a certain class in $\mathcal{C}^u$. Assuming $\mathcal{C}^l \ne \mathcal{C}^u$ and $\mathcal{C}^l \cap \mathcal{C}^u \ne \emptyset$, we define $\mathcal{C}^s=\mathcal{C}^l \cap \mathcal{C}^u$ as the set of known classes and $\mathcal{C}^n=\mathcal{C}^u \setminus \mathcal{C}^l$ as the set of unknown classes. $S=|\mathcal{C}^s|$ and $N=|\mathcal{C}^n|$ represent the number of known and unknown classes, respectively.
    
    In the previously mentioned methods related to open-world SSL, there is no concept of novel class in SSL, and it assumes $\mathcal{C}^l=\mathcal{C}^u$ by default, while NCD assumes that the unlabeled data does not contain any known classes, that is to say, $\mathcal{C}^l \cap \mathcal{C}^u=\emptyset$. Therefore, open-world SSL is inherently more challenging in nature.

\section{Method}
    In this section, we will first introduce the model architecture for self-supervised open-world classification. Then, we will formalize several learning objectives aimed at discovering multiple unknown classes. Finally, we will provide an overall algorithmic framework.
    \begin{figure*}[htbp]
      \centering
      \includegraphics[trim=0 0 0 0, clip, width=1.0\textwidth, ]{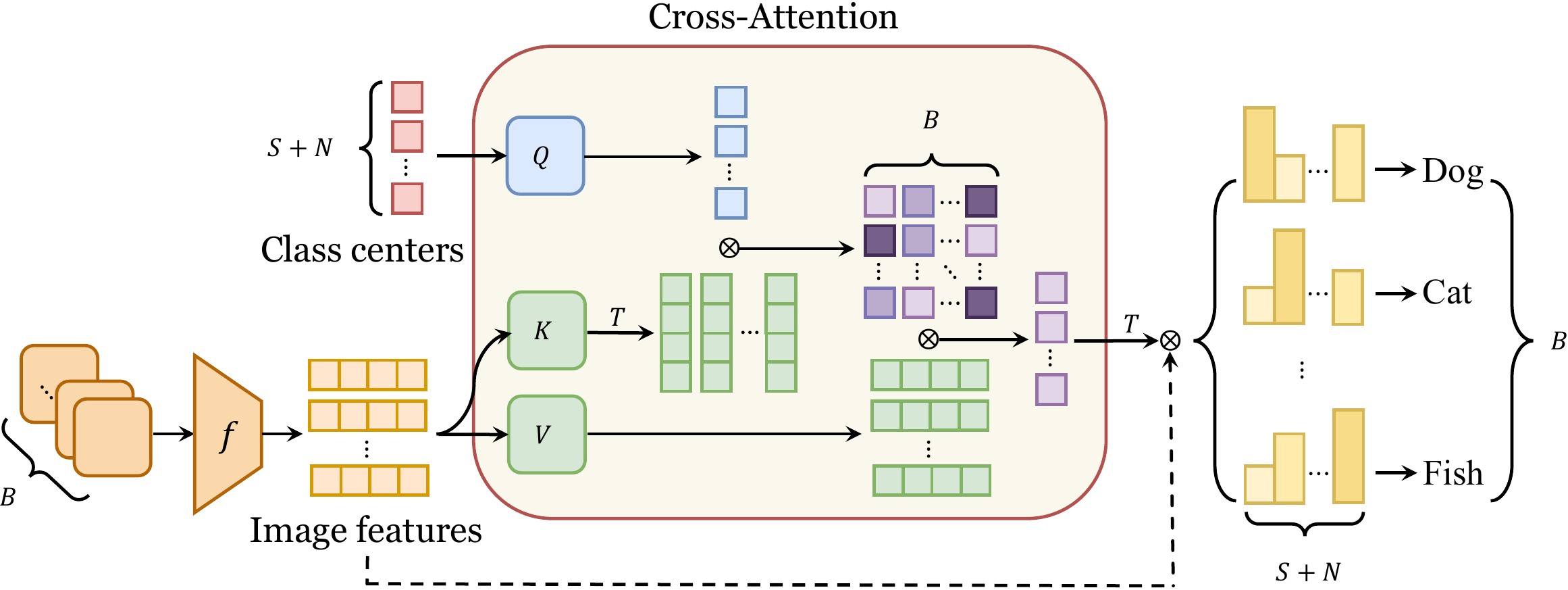}
      \caption{SSOC framework. SSOC models the concept of explicit classes and utilizes a cross-attention mechanism to facilitate dynamic interaction between image features and class centers. Through iterative steps, SSOC achieves self-learning of class centers. During inference, SSOC predicts the class based on the similarity between image features and class centers.}
      \label{fig:model}
    \end{figure*}
    \subsection{Self-Learning Open-World Classes}
        From the formalized problem definition, it can be understood that the key to the open-world SSL problem lies in effectively utilizing the unknown class samples in the unlabeled dataset. The main idea of SSOC is to explicitly self-learn open-world classes, regardless of whether they are known or unknown, which means we aim to learn representations of classes of the same dimensionality as image features. Obtaining class prototype representations alone is not a difficult problem; many previous unsupervised clustering methods can partition the data into several clusters and compute the center of each cluster. However, these methods do not utilize any labeled information and simply optimize the feature embedding model to achieve good clustering performance without incorporating class centers into the learning process. During the training process of open-world SSL, the data is input to the model in batches, and we expect the model to discover class information in each batch as much as possible and dynamically adjust the class centers. In backpropagation, explicit label information and the potential similarity relationships between samples are applied to optimize the class centers and the feature representations of images. To achieve this functionality, the core module of SSOC employs a cross-attention mechanism, which is the only parameterized network layer in SSOC apart from the feature extractor.

        The cross-attention mechanism is a method that captures the correlation features between two sequences. It was first introduced in Transformer \cite{DBLP:conf/nips/VaswaniSPUJGKP17} and is used to fuse the input sequence from the decoder with the output sequence from the encoder, obtaining encoder information relevant to the current position of the decoder. The cross-attention mechanism has significantly contributed to sequence modeling and natural language processing tasks. For example, it has been used in image-text classification to merge multi-modal input sequences \cite{DBLP:conf/iclr/JaegleBADIDKZBS22} and in machine translation to capture dependencies between distant positions in a sequence \cite{DBLP:conf/emnlp/Gheini0M21}.
        The internal structure of the cross-attention mechanism consists of three matrices: the query matrix $W^Q$, the key matrix $W^K$, and the value matrix $W^V$. The input sequences to the model can come from different modalities but must have the same dimensions. Using the query, key, and value matrices, the cross-attention mechanism calculates the correlation between sequences and fuses the value sequence relevant to the query sequence based on their weights. This mechanism allows the model to prioritize vital information in the sequences, enhancing task performance and capability.

        In the context of this paper, we assume that $X={\{x_i|x_i \in \mathbb{R}^D \}_{i=1}^B}$ represents the image data of the $t$-th batch, where $B$ represents the batch size. By utilizing a pre-trained deep neural network $f_{\theta}: \mathbb{R}^D \to \mathbb{R}^d$, we obtain the embedding features $\mathcal{Z}=\{z_i=f_{\theta}(x_i) | x_i \in X, z_i \in \mathbb{R}^d\}$. Here, $\theta$ denotes the model parameters, and $d$ represents the dimensionality of the embedding vectors. For ease of subsequent computations, SSOC sets the dimensions of $W^Q$, $W^K$, and $W^V$ as $d \times d$. We denote the class centers obtained from the $t$-th batch as $\mathcal{A}_{t}={\{a_i\}}_{i=1}^{S+N}$, particularly, $\mathcal{A}_0$ denotes the randomly initialized class center feature matrix, and $a_i$ represents the feature vector of the $i$-th class. In the cross-attention mechanism, we treat class features as query inputs, and data features as key and value inputs, then do matrix multiplication with the corresponding parameter matrix, namely $Q=\mathcal{A}_{t}W^Q, K=\mathcal{Z}W^K, V=\mathcal{Z}W^V$. The employed cross-attention mechanism can be expressed as follows.
        \begin{equation}\label{equ1}
        \Delta \mathcal{A}={\rm SoftMax}(\frac{QK^T}{\sqrt{d_k}})V
        \end{equation}
        The product of two vectors can usually represent the degree of similarity between them. In the equation above, $QK^T$ represents the attention matrix of size $(S+N) \times B$. The element at the $i$-th row and $j$-th column of this matrix can be interpreted as the correlation between the $i$-th class center and the $j$-th sample in the batch. A higher value indicates that the sample is likelier to belong to that class. The attention scores are then passed through a SoftMax layer and weighted sum with the data features, resulting in the cross-attention matrix $\Delta \mathcal{A}$ of size $(S+N) \times d$. Each row of $\Delta \mathcal{A}$ represents the weighted sum of the $i$-th class center and the data features, indicating the contribution of samples more similar to the $i$-th class in the batch. Conversely, samples less similar to the $i$-th class have smaller contributions. Therefore, $\Delta \mathcal{A}$ can be approximated as the class center feature vectors for the batch of samples. By computing the residual with the previous class centers, we can obtain the updated class centers $\mathcal{A}_{t+1}$.
        \begin{equation}\label{equ2}
         \mathcal{A}_{t+1}=\mathcal{A}_{t} + \Delta \mathcal{A}
        \end{equation}

        In this way, $\mathcal{A}_{t+1}$ retains the class information of most historical data while introducing the information of newly discovered classes in a residual manner. In the next iteration, it will be involved in computing the attention scores as class centers from the previous step. After several iterations, the class centers obtained from Equation 2 can represent the category concepts of all the data. Through this explicit learning approach, SSOC utilizes the cross-attention mechanism to achieve dynamic interaction between data features and class centers, enabling self-learning of categories. Compared to simply using clustering to obtain class centers, our method is less susceptible to the influence of data with extreme distribution deviations and better captures relevant features between samples and different classes.
        
        
        In the preceding sections, we obtained the residual representation of class centers, denoted as $\Delta \mathcal{A}$. Subsequently, SSOC employs a distance-based approach, utilizing dot product similarity to measure the distance between a sample and a class center within the same feature space for interpretable classification. The distances are then normalized into a probability distribution using the activation function. Through our experiments, we observed the challenge of achieving a balanced learning process between known and unknown classes, often resulting in slow learning of unknown classes and premature overfitting to known classes. To address this issue, we adjusted the temperature parameter of the SoftMax function when calculating the predicted probabilities for labeled data, while retaining the default value of 1 for unlabeled data. This adjustment involved utilizing a larger temperature parameter, leading to a flatter probability distribution, whereas a smaller value sharpens the output. By modifying the temperature parameter, our objective was to reduce the model's confidence in predicting known classes, thereby mitigating overfitting to known classes and enhancing the learning of unknown classes. Consequently, for a sample $x_i$, its probability distribution can be expressed as follows.
        \begin{equation}\label{equ11}
            p_i=\begin{cases}
                \sigma(z_i \cdot {\Delta \mathcal{A}}^T /\epsilon)& \text{$x_i \in \mathcal{D}^l$}\\ 
                \sigma(z_i \cdot {\Delta \mathcal{A}}^T)  & \text{Others.} \end{cases}
        \end{equation}
        where $\sigma$ represents the SoftMax optimization operator, $\epsilon$ is the scaling hyperparameter, and $p_{ij}$ represents the predicted probability of sample $x_i$ belonging to class $j$. In the subsequent discussions, we employ $\hat{p}_i=\max_j{p_{i}}$ to denote the maximum confidence score and $\hat{y}_i=\arg \max p_i$ to represent the pseudo-label of the sample, which is then converted into one-hot form. Lastly, we illustrate the architecture of SSOC in Fig. \ref{fig:model}.

    \subsection{Optimize Objective}
        To assist SSOC in learning unknown classes in the open-world scenario, we designed a set of optimization objectives. Specifically, our loss function consists of three components: a cross-entropy loss that selects confident unlabeled data, a pairwise similarity loss that selects confident pairs of related samples, and a regularization term that prevents overfitting to known classes. The overall objective is as follows.
        \begin{equation}\label{equ4}
            \mathcal{L}=\mathcal{L}_{CE}+\alpha \mathcal{L}_{BCE}+\beta \mathcal{L}_{RE}
            \end{equation}
        where $\alpha$ and $\beta$ are balancing hyper-parameters. Next, we will explain these three loss objectives in detail.

        \noindent \emph{Cross-Entropy.} Cross-entropy is a metric widely used to measure the difference between two probability distributions, commonly applied in classification tasks and probability estimation. For labeled data, SSOC maximizes the utilization of label information by minimizing the cross-entropy loss. The ground-truth labels of the labeled data $\mathcal{D}^l$ are represented as one-hot vectors $Y^l$, and the probability distribution $P^l$ of the labeled data is calculated using Eq. \ref{equ11}. The supervised loss term can be expressed as follows.
        \begin{equation}\label{equ5}
           \mathcal{L}_{CE}^l=-\frac{1}{\mathcal{M}} \sum_{i=1}^{\mathcal{M}} \sum_{j=1}^{S+N} Y^l_{ij}\log{P^l_{ij}}
            \end{equation}
        For unlabeled data, they lack the necessary ground-truth labels, and direct computation of cross-entropy is impossible. To address this issue, many SSL methods employ pseudo-labels to compute the pseudo-supervised loss for unlabeled samples, helping the model extract information from unlabeled data. However, in many scenarios, noisy data in the dataset can result in incorrect or unreliable pseudo-labels, which can interfere with the model and degrade performance. To reduce noise interference and ensure the robustness of the model, we use a threshold to filter out high-confidence pseudo-labels for model training. Specifically, by calculating the probability distribution $P^u$ and pseudo-labels $\hat{Y}^u$ for the unlabeled sample set $\mathcal{D}^u$ using Eq. \ref{equ11}, we only consider unlabeled samples with maximum confidence scores higher than the threshold $\tau_1$ for computing the cross-entropy loss.
        \begin{equation}\label{equ6}
            \mathcal{L}_{CE}^u=-\frac{1}{\mathcal{N}} \sum_{i=1}^{\mathcal{N}}{\mathbb I({\hat P}_i^u > \tau_1) \log{{\hat P}_i^u}}
            \end{equation}
        where $\mathbb{I}(\cdot)$ is the indicator function. In summary, the overall cross-entropy loss of SSOC is the weighted sum of the two components above.
        \begin{equation}\label{equ7}
            \mathcal{L}_{CE}=\gamma \mathcal{L}_{CE}^l+\delta \mathcal{L}_{CE}^u
            \end{equation}
        where $\gamma$ and $\delta$ are balancing hyper-parameters.
        
        In multiple experiments, we observed that during the initial training stages, the model has limited knowledge about unknown classes and exhibits higher uncertainty when predicting unlabeled data. As a result, the model initially selects a small number of unknown class samples, which is insufficient to develop a comprehensive understanding of unknown classes. However, the labeled data has precise and unambiguous ground-truth labels, and the supervision from $\mathcal{L}_{CE}^l$ is relatively strong. This can cause the model to be biased towards known classes and incorrectly classify unknown class samples as single known classes. To mitigate this issue and enhance the model's attention towards unknown classes, we introduce perturbations to the unlabeled data. For each unlabeled image sample, we generate two augmented images, $r_1$ and $r_2$, and concatenate their feature vectors with those of the labeled data. These concatenated feature vectors are collectively used to compute the class center increments in Eq. \ref{equ1}. When calculating the pseudo-supervised loss, we retain only the image data of $r_1$, treating the pseudo-labels of $r_2$ as the pseudo-labels of $r_1$. This approach enables the model to learn more invariant features of unknown classes.

        \noindent \emph{Pairwise Similarity.} Binary cross-entropy (BCE) loss is commonly used for binary classification tasks. To enable SSOC to learn better category features, we employ BCE loss to constrain the similarity between sample pairs. This idea has been successfully implemented in previous work such as NCD \cite{DBLP:conf/cvpr/ZhongFRL0S21}, \cite{DBLP:conf/iclr/HanREVZ20}. In the embedding space, there are only two possible relationships between samples: same or different classes. The BCE loss aims to bring similar samples closer and push dissimilar samples apart. For labeled data, we directly use the ground-truth labels to determine whether they belong to the same class. For unlabeled data, we measure the similarity between sample pairs using the cosine similarity of their feature embeddings. To mitigate the negative impact of unreliable noise samples, we introduce a threshold $\tau_2$ to filter out sample pairs with sufficient confidence.

        \begin{equation}\label{equ8}
        \begin{aligned}
         \mathcal{L}_{BCE} &= -\sum_{i=1}^{\mathcal{M}+\mathcal{N}}{\sum_{j=1}^{\mathcal{M}+\mathcal{N}}{\mathbb{I}(\min({\hat P}_i, {\hat P}_j)>\tau_2)}} \\
          & \quad \quad \times [S_{ij}\log P_i^T P_j + (1-S_{ij})\log (1-P_i^T P_j)]
        \end{aligned}
        \end{equation}
        
        \begin{equation}\label{equ9}
            S_{ij}=\begin{cases}\mathbb{I}(y_i=y_j) & \text{$x_i,x_j \in \mathcal{D}^l$}\\\cos (\mathcal{Z}_i, \mathcal{Z}_j)  & \text{Others.}\end{cases}
            \end{equation}
        where $P=P^l \cup P^u$ represents the probability distribution of all data, and $\mathcal{Z}=\mathcal{Z}^l \cup \mathcal{Z}^u$ represents the features of all data.

        During the minimization of the BCE loss, sample pairs with highly similar features (i.e., $s_{ij}$ close to 1) have their probability distributions optimized to be more similar. Conversely, sample pairs with larger feature differences have their probability distributions optimized to be more distinct. Therefore, in our approach, the BCE loss aligns the prediction space with the embedding space, aiding the model in learning inter-class differences and intra-class similarities.
        
        \noindent \emph{Maximum Entropy Regularization.} In our experiments, we observed that during the initial training stage of SSOC, the cross-entropy loss dominated, resulting in the clustering of class centroids and impeding their separation. As a result, all unlabeled data might be mistakenly classified into a single category, which is undesirable. To promote a more uniform distribution of predicted classes, we introduced a maximum entropy regularization term to increase the uncertainty of the model's predictions. Maximum entropy regularization is essentially the empirical entropy, which estimates the uncertainty of the prior probability distribution based on the observed data frequencies. It can be expressed as follows.
        \begin{equation}\label{equ10}
            \mathcal{L}_{RE}=\sum_{i=1}^{\mathcal{M}+\mathcal{N}}p_i \log p_i
            \end{equation}
        In SSOC, we apply the above equation to all samples, maximizing the empirical entropy to make the model's predicted distribution more flexible and diverse, thus providing more opportunities for assigning unlabeled data to different classes. Experimental results demonstrate that the maximum entropy regularization term effectively enhances the model's robustness on unknown classes. In Algorithm 1, we provide a detailed description of the training process of SSOC.

    \renewcommand{\algorithmicrequire}{\textbf{Input:}}  
    \renewcommand{\algorithmicensure}{\textbf{Output:}} 
    \begin{algorithm}[h]
      \caption{Robust Semi-Supervised Learning for Self-learning Open-World Classes} 
      \begin{algorithmic}[1]
        \Require
            Labeled datasets $\mathcal{D}^l={\{(x_i, y_i)\}}_{i=1}^\mathcal{M}$, unlabeled dataset 
            $\mathcal{D}^u={\{x_i\}}_{i=1}^\mathcal{N}$, the seen classes number $S$, the novel classes 
            number $N$, the number of iterations for updating class centers $t$, the current class centers $\mathcal{A}_{t}={\{a_i\}}_{i=1}^{S+N}$, 
            a pre-trained backbone $f_\theta$, the initialized cross-attention layers 
            $g_\varphi$, the number of the training epochs $T$, the mini-batch number $B$.
        \Ensure
           A fine-tuned model $f_{\theta}$ and the trained cross-attention layers $g_{\varphi}$.
        \For{epoch from 1 to $T$}
            \For{batch from 1 to $B$}
                \State $X^l, Y^l \gets$ Sample mini-batch from $\mathcal{D}^l$
                \State $X^u_1, X^u_2 \gets$ Sample and augment mini-batch from $\mathcal{D}^u$
                \State $\mathcal{Z}^l, \mathcal{Z}^u_1, \mathcal{Z}^{u}_2 \gets f_{\theta}(X^l), f_{\theta}(X^u_1), f_{\theta}(X^{u}_2)$
                \State $\Delta \mathcal{A} \gets g_\varphi(\mathcal{Z}^l \cup \mathcal{Z}^u_1 \cup \mathcal{Z}^u_2, \mathcal{A}_t)$ with (Eq. \ref{equ1})
                \State Compute $P^l, P^u_1, P^u_2$ with (Eq. \ref{equ11})
                \State $\hat{Y}^u_2 \gets \arg\max(P^u_2)$
                \State $\mathcal{Z}, P, Y \gets \mathcal{Z}^l \cup \mathcal{Z}^u_1, P^l \cup P^u_1, Y^l \cup \hat{Y}^{u}_2$
                \State Compute $\mathcal{L}_{CE}$ with (Eq. \ref{equ7})
                \State Compute $\mathcal{L}_{BCE}$ with (Eq. \ref{equ8})
                \State Compute $\mathcal{L}_{RE}$ with (Eq. \ref{equ10})
                \State Compute updated class centers $\mathcal{A}_{t+1}$ with (Eq. \ref{equ2})
                \State $f_\theta, g_\varphi \gets $ Optimize with (Eq. \ref{equ4})
            \EndFor
        \EndFor
      \end{algorithmic}
    \end{algorithm}

\section{Experiment}
    This section describes the experimental setup for SSOC, including datasets, comparison methods, implementation details, and evaluation metrics. We present experimental results on various data splits, validate loss function effectiveness through ablation study, and assess parameter impact.
    
    \subsection{Experimental Setup}
        \noindent \emph{Dataset.} To validate the effectiveness of SSOC, we conducted experiments on three widely used computer vision benchmark datasets: CIFAR-10 \cite{krizhevsky2009learning}, CIFAR-100 \cite{krizhevsky2009learning}, and ImageNet-100 \cite{DBLP:journals/ijcv/RussakovskyDSKS15}. The CIFAR-10 and CIFAR-100 datasets consist of 60,000 images with a resolution of $32 \times 32$. Among them, 50,000 images are used for training and 10,000 for testing. CIFAR-10 consists of 10 classes, with approximately 6,000 images per class, while CIFAR-100 contains 100 classes, with around 600 images per class. ImageNet-100 is a subset of the ILSVRC2012 dataset, consisting of 100 classes selected from the original 1,000 classes \cite{van2020scan}. For consistent comparison with other studies, we used the same 100 classes as ORCA and NACH. In all datasets, we applied random cropping and rotation as data augmentation techniques to augment the dataset. In the main experiments, we treated the first 50\% of classes in each dataset as known classes and the remaining as unknown classes. We used 50\% of the labeled data from the known classes as labeled data, while the rest, along with the data from the unknown classes, were treated as unlabeled data. Additionally, we presented results for different label ratios and novel ratios. Random data splits were used in all experiments to ensure the generalizability of the results.

        \noindent \emph{Comparison method.} We compare SSOC with SSL, open-set SSL, NCD, and existing open-world SSL methods. SSL and open-set SSL methods are limited to classifying known classes. To extend them to open-world SSL scenarios, K-means clustering of unknown classes is performed, allowing evaluation of their performance on novel classes. FixMatch \cite{DBLP:conf/nips/SohnBCZZRCKL20} is chosen as the representative SSL method, and $\rm DS^3L$ \cite{guo2020safe} and CGDL \cite{DBLP:conf/cvpr/SunYZLP20} as representative open-set SSL methods. Since SSL methods do not include the concept of novel classes, samples with low confidence in SoftMax outputs are treated as unknown classes. For NCD-based methods, DTC \cite{DBLP:conf/iccv/HanVZ19} and RankStats \cite{DBLP:conf/iclr/HanREVZ20} are selected for comparison. Since they can only cluster unknown classes and lack performance on known classes, we use the Hungarian algorithm to perform maximum weighted matching between the known classes in the labeled data and the clustered classes \cite{kuhn1955hungarian}, \cite{DBLP:conf/iclr/CaoBL22}, \cite{DBLP:conf/nips/GuoZWSL22}, and evaluate the results separately for these known classes. For open-world SSL, we compare the reported performances of ORCA and NACH, the two existing methods in this research area, to the best of our knowledge.

        \noindent \emph{Implementation Details.} For the CIFAR-10 dataset, we employ ResNet34 \cite{he2016deep} as the backbone network and use two Adam optimizers to optimize the backbone network and the cross-attention layers separately. We fine-tune the backbone network with a lower learning rate of 1e-4, while the cross-attention layers are trained with a slightly higher learning rate of 5e-3, focusing on learning category information. Both optimizers have momentum parameters set to $(0.9, 0.99)$. We train the models with a batch size of 128 for 200 epochs. For the CIFAR-100 dataset, we use ResNet18 \cite{he2016deep} as the backbone network. The backbone network and the cross-attention layers are trained with a learning rate of 1e-4. The batch size is 512, and the models are trained for 500 epochs. For the ImageNet-100 dataset, we adopt ResNet50 \cite{he2016deep} as the backbone network. The backbone network and the cross-attention layers are trained with learning rates of 1e-5 and 3e-4, respectively. The models are trained with a batch size of 100 for 200 epochs. In all experiments, we employ an early stopping strategy and dynamically adjust the learning rate using the cosine annealing method. The experiments on the CIFAR dataset are conducted on 8 V100 GPUs, while the experiments on ImageNet-100 are performed using 4 NVIDIA 3090 GPUs.
        
        We utilize a pre-trained ResNet model to extract high-quality image features. In our experiments, we extract the initial embedding vectors $\mathcal{Z}^l \cup \mathcal{Z}^u$ of all the data using the pre-trained backbone before training. Subsequently, we apply the K-means++ algorithm \cite{bahmani2012scalable} to these vectors for unsupervised clustering, resulting in the initialization of class center representations $\mathcal{A}_0$. This initialization, which incorporates prior knowledge, facilitates the learning process of the model. Furthermore, in practical scenarios, it might be infeasible to obtain samples for unknown classes, we can resort to random initialization of prototypes for the unknown classes, while employing the clustering method mentioned earlier to obtain prototypes for the known classes. The code for SSOC can be found at \href{https://github.com/njustkmg/SSOC}{https://github.com/njustkmg/SSOC}.
        
        \noindent \emph{Evaluation Metrics.} We adopt the evaluation approach used in \cite{DBLP:conf/iclr/CaoBL22}, \cite{DBLP:conf/nips/GuoZWSL22} and report the accuracy of SSOC on known classes, unknown classes, and all classes. Additionally, in the ablation experiments, we report the normalized mutual information (NMI) on the novel classes. It is worth noting that since the learned concepts of novel classes by the model are unordered, before computing the accuracy on unknown classes and all classes, we utilize the Hungarian maximum weighted matching algorithm \cite{kuhn1955hungarian} for label alignment to obtain the optimal matching between the clustering labels of unknown classes and the ground-truth labels.

    \subsection{Main Results}
        \begin{table*}
        \centering
        \renewcommand{\arraystretch}{1.5}
        \caption{Accuracy on CIFAR-10, CIFAR-100, and ImageNet-100, with 50\% label ratio and 50\% novel ratio. The best results are highlighted in bold.}
        \label{tab1}
            \begin{tabular}{c|c|ccc|ccc|ccc}
                \hline \hline
                \multirow{2}{*}{Method Classes} &{Dataset} &\multicolumn{3}{c|}{{CIFAR-10}} &\multicolumn{3}{c|}{{CIFAR-100}} &\multicolumn{3}{c}{{ImageNet-100}} \\
                \cline{2-11}
                &{Method} &{Seen} &{Novel} &{All} &{Seen} &{Novel} &{All} &{Seen} &{Novel} &{All} \\
                \hline
                SSL &{FixMatch} &{71.5} &{50.4} &{49.5} &{39.6} &{23.5} &{20.3} &{65.8} &{36.7} &{34.9} \\
                \hline
                {} &{$\rm DS^3L$} &{77.6} &{45.3} &{40.2} &{55.1} &{23.7} &{24.0} &{71.2} &{32.5} &{30.8} \\
                \multirow{-2}{*}{{Open-set SSL}} &{CGDL} &{72.3} &{44.6} &{39.7} &{49.3} &{22.5} &{23.5} &{67.3} &{33.8} &{31.9} \\
                \hline
                {} &{DTC} &{53.9} &{39.5} &{38.3} &{31.3} &{22.9} &{18.3} &{25.6} &{20.8} &{21.3} \\
                \multirow{-2}{*}{{NCD}} &{RankStats} &{86.6} &{81.0} &{82.9} &{36.4} &{28.4} &{23.1} &{47.3} &{28.7} &{40.3} \\
                \hline
                {} &{ORCA} &{88.2} &{90.4} &{89.7} &{66.9} &{43.0} &{48.1} &{89.1} &{72.1} &{77.8} \\
                {} &{NACH} &{89.5} &{92.2} &{91.3} &{68.7} &{47.0} &{52.1} &{91.0} &{75.5} &{79.6} \\
                \multirow{-3}{*}{{Open-world SSL}} &{SSOC} &{\textbf{93.3}} &{\textbf{92.6}} &{\textbf{92.8}} &{\textbf{69.0}} &{\textbf{48.0}} &{\textbf{53.1}} &{\textbf{91.4}} &{\textbf{78.4}} &{\textbf{82.7}} \\
                \hline \hline
            \end{tabular}
        \end{table*}
        
        \noindent \emph{Comparison of main results.} In Tab. (\ref{tab1}), we display SSOC's classification accuracies alongside other methods on CIFAR-10, CIFAR-100, and ImageNet-100 with 50\% label ratio and 50\% novel ratio. SSOC consistently outperforms SSL, open-set SSL, and NCD methods in open-world SSL scenarios, achieving superior accuracy for known, unknown, and all classes across all datasets. Compared to the best-performing method, RankStats, we achieved improvements of 30\% and 42.4\% on all classes of CIFAR-100 and ImageNet-100, respectively. Furthermore, SSOC outperforms the two open-world SSL methods overall. Compared to NACH, we achieved a 3.8\% improvement in accuracy for known classes in CIFAR-10, and significantly improved the performance for unknown classes in the challenging ImageNet-100 dataset, with accuracy improvements of 2.9\% and 3.1\%, respectively. Our experimental results demonstrate that SSOC effectively addresses the open-world SSL problem.
        
        \begin{table*}
        \renewcommand{\arraystretch}{1.5}
        \centering
        \caption{Accuracy on CIFAR-10, CIFAR-100, and ImageNet-100 when changing the label ratio, with 50\% novel ratio. The best results are highlighted in bold.}
        \label{tab2}
        \setlength{\tabcolsep}{4mm}{
            \begin{tabular}{c|ccc|ccc|ccc}
                \hline \hline
                Dataset &\multicolumn{3}{c|}{CIFAR-10} &\multicolumn{3}{c|}{CIFAR-100} &\multicolumn{3}{c}{ImageNet-100} \\
                \hline
                Label Ratio &ORCA &NACH &{SSOC} &{ORCA} &{NACH} &{SSOC} &{ORCA} &{NACH} &{SSOC} \\
                \hline
                {10\%} &{84.10} &{88.10} &\textbf{90.90} &{38.60} &{43.38} &\textbf{43.82} &{69.70} &{66.81} &\textbf{75.47} \\
                \hline
                {30\%} &{87.70} &{90.21} &\textbf{92.24} &{43.71} &\textbf{51.06} &{50.66} &{70.79} &{70.74} &\textbf{76.83} \\
                \hline
                {50\%} &{89.70} &{91.30} &\textbf{92.79} &{48.10} &{52.10} &\textbf{53.13} &{77.80} &{79.60} &\textbf{82.69} \\
                \hline \hline
            \end{tabular}}
        \end{table*}
        \noindent \emph{Change the label ratio.} To demonstrate the effectiveness of SSOC in scenarios with a limited amount of labeled data, we fixed the novel ratio at 50\% and compared the performance of ORCA, NACH, and SSOC with label ratios of 10\% and 30\%. Tab. \ref{tab2} presents the accuracy for all classes, including results from the ORCA and NACH papers. It can be observed that as the labeled data decreases, the performance of all three methods declines. Notably, on CIFAR-10 and ImageNet-100, SSOC only incurs performance drops of 1.89\% and 7.22\%, respectively, when the label ratio decreases from 50\% to 10\%, while NACH experiences declines of 3.2\% and 12.79\%, respectively. Additionally, on the ImageNet-100 dataset, when the label ratios are 10\% and 30\%, SSOC achieves an overall accuracy that is 8.66\% and 6.09\% higher than NACH, respectively. Though our CIFAR-100 result with a 30\% label ratio is slightly inferior to NACH, it remains 6.95\% above ORCA. Overall, SSOC demonstrates strong robustness and performs well in situations with limited labeled data.

        \begin{table*}
        \centering
        \renewcommand{\arraystretch}{1.5}
        \caption{Accuracy on CIFAR-10, CIFAR-100, and ImageNet-100 when changing the novel ratio, with 50\% label ratio. The best results are highlighted in bold.}
        \label{tab3}
        \setlength{\tabcolsep}{4mm}{
        \begin{tabular}{c|ccc|ccc|ccc}
            \hline \hline
            {Dataset} &\multicolumn{3}{c|}{{CIFAR-10}} &\multicolumn{3}{c|}{{CIFAR-100}} &\multicolumn{3}{c}{{ImageNet-100}} \\
            \hline
            {Novel Ratio} &{ORCA} &{NACH} &{SSOC} &{ORCA} &{NACH} &{SSOC} &{ORCA} &{NACH} &{SSOC} \\
            \hline
            $10\%$ &91.20 &\textbf{95.84} &95.14 &61.35 &\textbf{66.77} &64.29 &90.36 &89.34 &\textbf{91.58} \\
            \hline
            $30\%$ &91.08 &93.08 &\textbf{95.13} &52.17 &54.12 &\textbf{56.39} &86.09 &80.03 &\textbf{90.49} \\
            \hline
            $50\%$ &89.70 &91.30 &\textbf{92.79} &48.10 &52.10 &\textbf{53.13} &77.80 &79.60 &\textbf{82.69} \\
            \hline
            $70\%$ &81.52 &86.44 &\textbf{87.38} &38.12 &43.93 &\textbf{44.78} &57.57 &53.88 &\textbf{73.58} \\
            \hline
            $90\%$ &56.63 &59.19 &\textbf{75.52} &28.29 &26.79 &\textbf{37.40} &41.10 &43.56 &\textbf{65.86} \\
            \hline \hline
            \end{tabular}}
        \end{table*}
        \noindent \emph{Change the novel ratio.} By fixing the label ratio at 50\%, we explored the impact of varying novel ratios on open-world SSL methods. Tab. \ref{tab3} shows the overall class accuracy of ORCA, NACH, and SSOC at novel ratios of 10\%, 30\%, 70\% and 90\%. Since the ORCA and NACH papers lack these experimental data, we reproduced their open-source code. Observing the data, it is evident that as the number of unknown classes increases, the performance of all three methods decreases, but SSOC exhibits a smaller performance drop than the other two methods. Notably, SSOC achieves impressive results in scenarios with a higher novel ratio. When the novel ratio is 90\%, SSOC achieves accuracy improvements of 10.61-22.3\% compared to NACH on the three datasets, and a 16.01\% accuracy improvement compared to ORCA on the ImageNet-100 dataset with a novel ratio of 70\%. Furthermore, on the CIFAR dataset with a novel ratio of 10\%, SSOC did not surpass NACH, indicating that NACH focuses more on classifying known classes, while SSOC places greater emphasis on discovering unknown classes and still achieves excellent results even with a small number of known classes.
        
        Through these three experiments, we have demonstrated the effectiveness of SSOC and its outstanding robustness and generalization performance. SSOC outperforms ORCA and NACH in handling scenarios with limited labeled data or many novel classes, making it applicable in a wider range of scenarios and having significant practical implications.

    \subsection{Ablation Study}
        To validate the effectiveness of different loss functions, we conducted ablation experiments on ImageNet-100 with a label ratio and novel ratio of 50\%. Tab. \ref{tab4} reports the accuracy on known classes, unknown classes and the overall accuracy, as well as the NMI on unknown classes. In the first three rows, we removed the cross-entropy loss, pairwise similarity loss, and maximum entropy regularization term, respectively, and used the remaining loss as the objective function. The experimental results show that $\mathcal{L}_{CE}$ plays a crucial role in SSOC. The model relies on the supervised loss from labeled data to obtain essential ground-truth class information and learns about unknown classes through pseudo-supervised loss. Furthermore, in the experiment where $\mathcal{L}_{RE}$ was removed, the performance on novel classes significantly deteriorated, confirming the beneficial effect of maximum entropy regularization on learning novel classes. 

        \begin{table}
        \renewcommand{\arraystretch}{1.3}
        \caption{Ablation study on ImageNet-100 dataset, we report the accuracy and NMI on unknown classes.}
        \label{tab4}
        \setlength{\tabcolsep}{4mm}{
        \begin{tabular}{c|cccc}
        \hline \hline
        \textbf{Approach} & \textbf{Seen} & \textbf{Novel} & \textbf{Novel(NMI)} & \textbf{All} \\ \hline
        \pmb{w/o} $\mathcal{L}_{CE}$ & 2.06 & 2.03 & 3.67 & 1.35 \\
        \pmb{w/o} $\mathcal{L}_{BCE}$ & 90.60 & 75.51 & 75.29 & 80.33 \\
        \pmb{w/o} $\mathcal{L}_{RE}$ & 81.48 & 37.77 & 63.88 & 47.00 \\
        \textbf{SSOC} & 91.39 & 78.37 & 78.32 & 82.69 \\ \hline \hline
        \end{tabular}}
        \vspace{-1.0em}

        \end{table}

    \subsection{Parametric Analysis}
        \begin{figure*}[ht]
            \centering
            \begin{minipage}[t]{0.29\linewidth}
                \includegraphics[width=\textwidth]{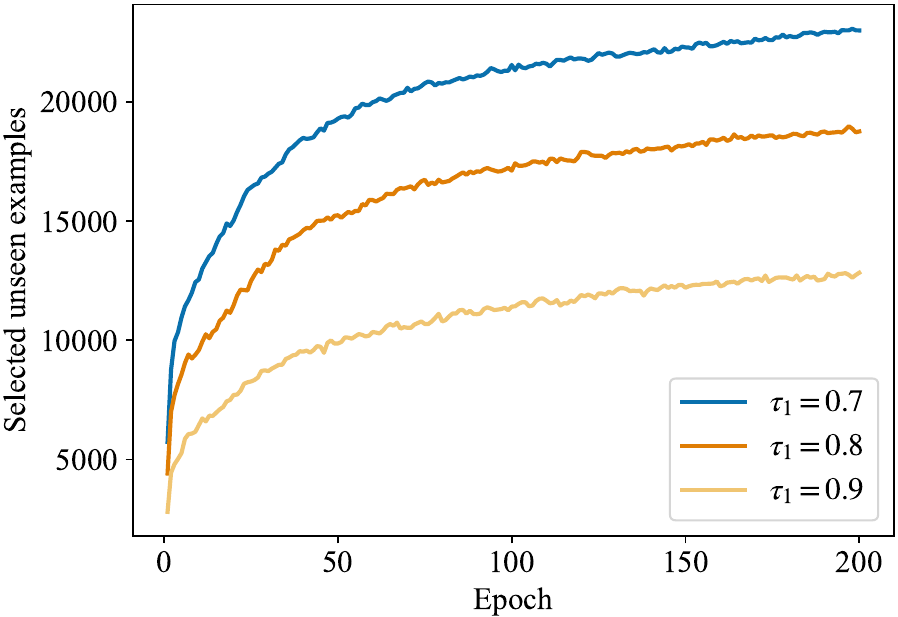}
                \parbox{\textwidth}{\centering ({\it a}) {Selected pseudo-labels for unseen classes (different $\tau_1$)}}
            \end{minipage}
            \begin{minipage}[t]{0.29\linewidth}
                \includegraphics[width=\textwidth]{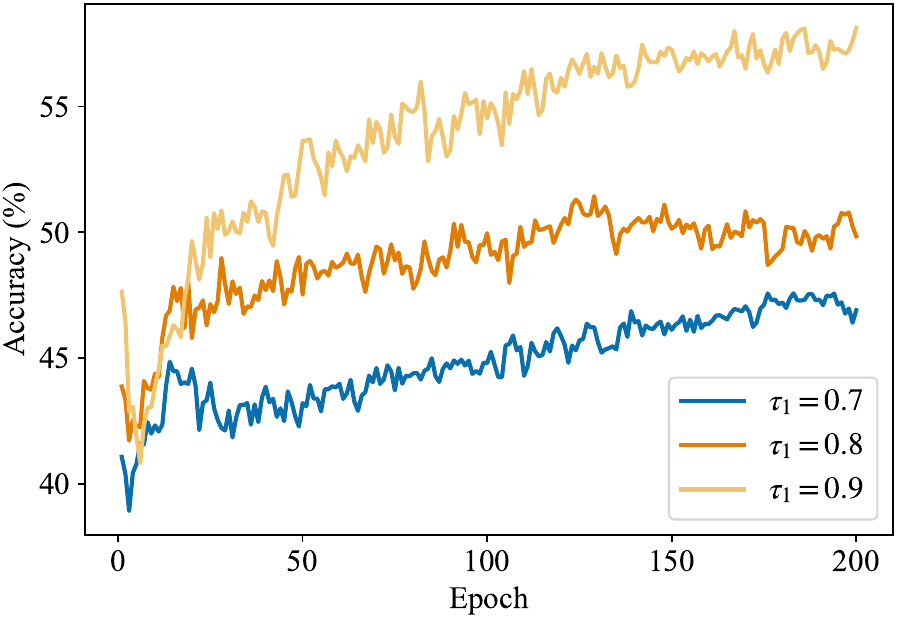}
                \parbox{\textwidth}{\centering ({\it b}) {Accuracy of pseudo-labels for unseen classes (different $\tau_1$)}}
            \end{minipage}
            \begin{minipage}[t]{0.29\linewidth}
                \includegraphics[width=\textwidth]{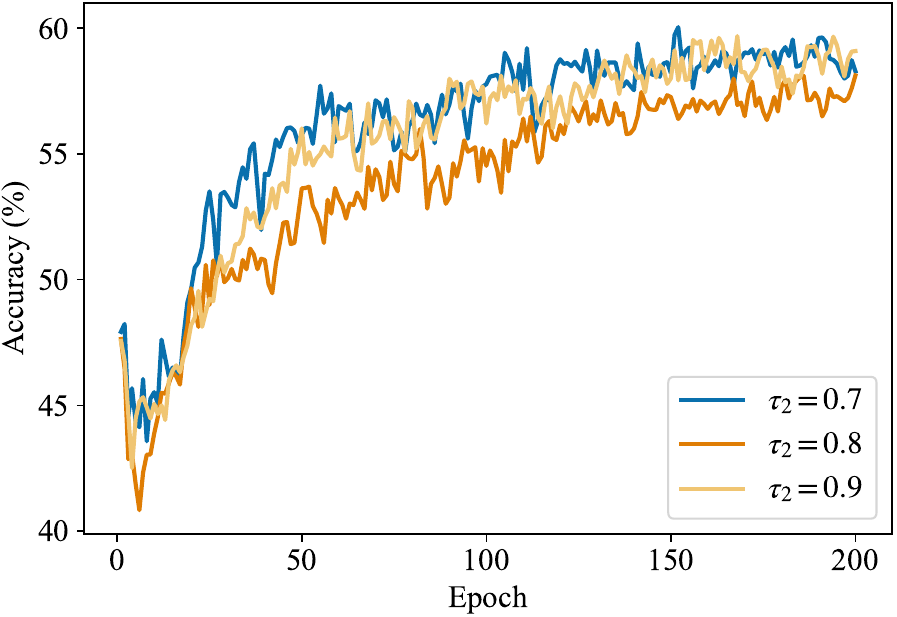}
                \parbox{\textwidth}{\centering ({\it c}) {Accuracy of pseudo-labels for unseen classes (different $\tau_2$)}}
            \end{minipage}
            \caption{Analysis of threshold effect on the CIFAR-100 dataset. Fig. \ref{fig:para1}(a) and Fig. \ref{fig:para1}(b) illustrate the number of selected pseudo-labels and the corresponding accuracy for different values of $\tau_1$. Fig. \ref{fig:para1}(c) presents the accuracy of the selected pseudo-labels for different values of $\tau_2$.}
            \label{fig:para1}
        \end{figure*}
        \noindent \emph{Effect of threshold selection.} To analyze the impact of thresholds on the experimental results, we conducted experiments on the CIFAR-100 dataset with different values of $\tau_1$ and $\tau_2$, with a label ratio and novel ratio of 50\%. In Fig. \ref{fig:para1} (a) and (b), we varied the value of $\tau_1$ and presented the number of unlabeled samples selected by the threshold during each epoch of training in (a), and the pseudo-label accuracy of selected unknown class samples in (b). It can be observed that a lower $\tau_1$ is insufficient to filter out falsely labeled samples, which can interfere with the model's classification ability. On the other hand, a larger $\tau_1$ excessively eliminates unlabeled data, failing to provide sufficient unknown class samples for the model to learn, resulting in poor performance on unknown classes. In (c), we provide the pseudo-label accuracy of selected unknown class samples for different $\tau_2$ values. In all CIFAR-100 experiments, we set $\tau_1$ to 0.6 and $\tau_2$ to 0.8.

        \begin{figure*}[ht]
            \centering
            \begin{minipage}[t]{0.29\linewidth}
                \includegraphics[width=\textwidth]{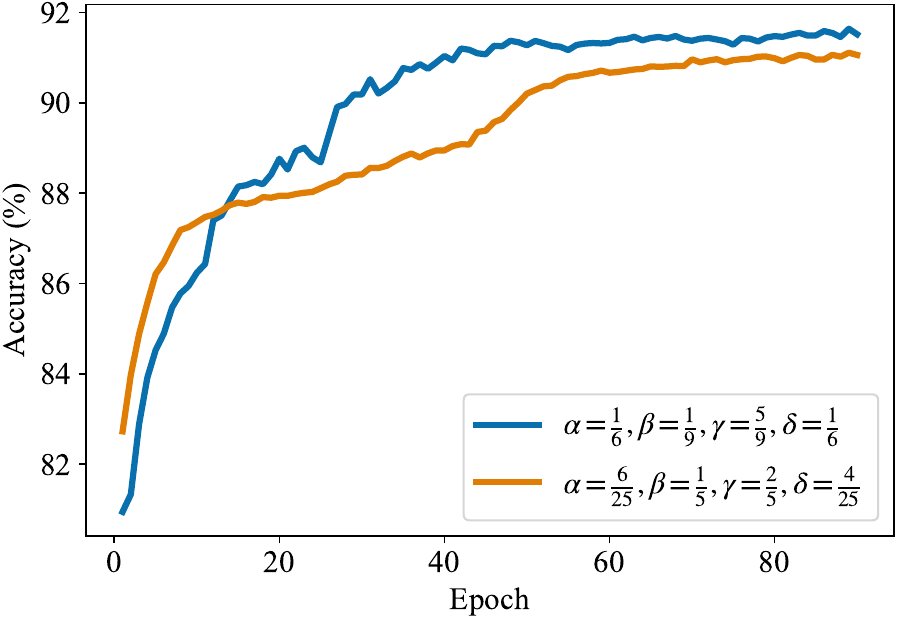}
                \parbox{\textwidth}{\centering ({\it a}) {Accuracy on seen classes}}
            \end{minipage}
            \begin{minipage}[t]{0.29\linewidth}
                \includegraphics[width=\textwidth]{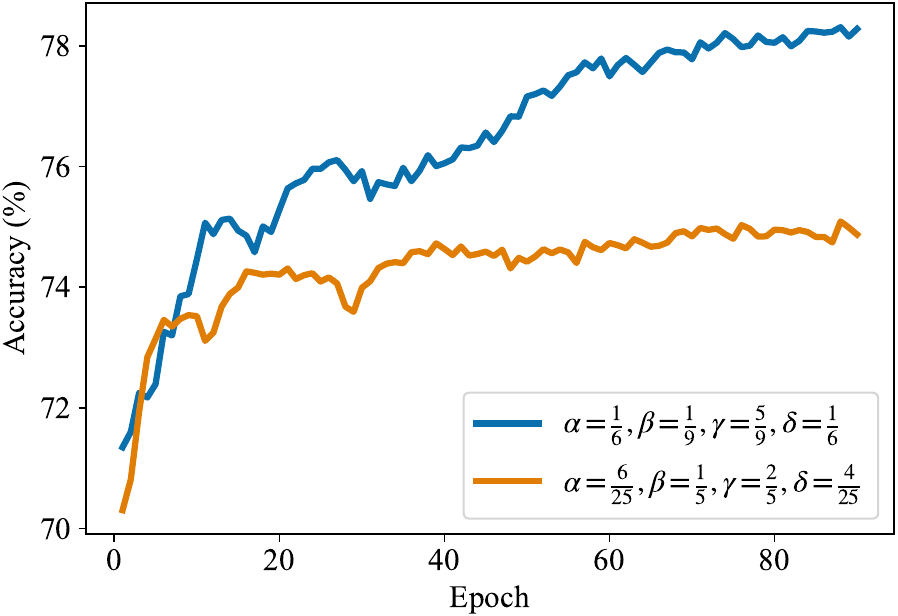}
                \parbox{\textwidth}{\centering ({\it b}) {Accuracy on unseen classes}}
            \end{minipage}
            \begin{minipage}[t]{0.29\linewidth}
                \includegraphics[width=\textwidth]{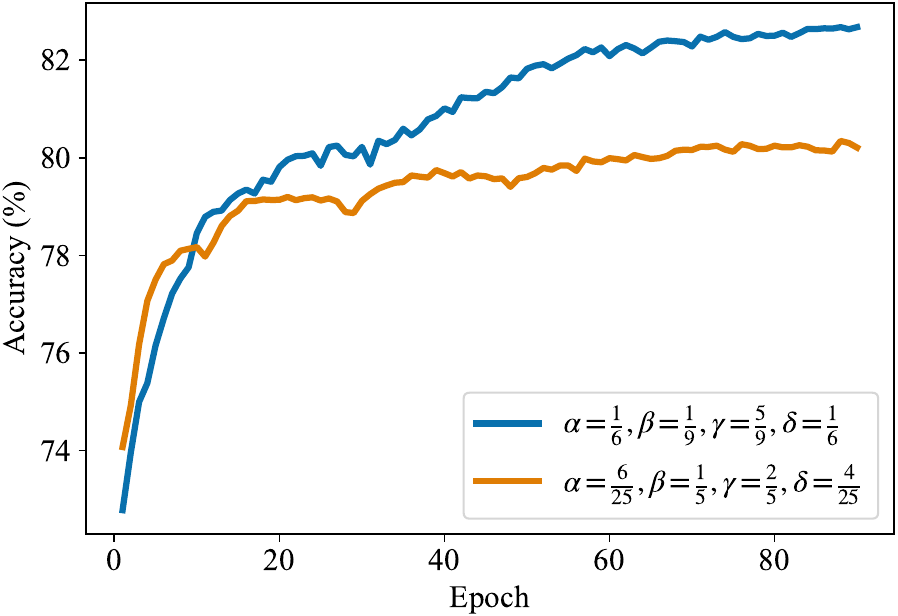}
                \parbox{\textwidth}{\centering ({\it c}) {Accuracy on overall classes}}
            \end{minipage}
            \caption{Analysis of the effect of loss balance hyper-parameters on the ImageNet-100 dataset. We select two sets of parameter configurations, and in Fig. \ref{fig:para2}(a), Fig. \ref{fig:para2}(b), and Fig. \ref{fig:para2}(c), we present the accuracy on seen classes, unseen classes, and overall classes, respectively.}
            \label{fig:para2}
        \vspace{-1.5em}
        \end{figure*}

        \noindent \emph{Effect of the balance hyper-parameter on the loss.} To investigate the impact of different loss weights on the results, we present the accuracies of SSOC on the ImageNet-100 dataset for two sets of loss weights, in Fig. \ref{fig:para2}. It can be observed that elevating the weight of loss terms associated with unknown classes seems to constrain the model's capacity to learn about unknown classes, which subsequently impacts the overall performance. This suggests that the loss terms $\mathcal{L}_{CE}^u$, $\mathcal{L}_{BCE}$, and $\mathcal{L}_{RE}$ are closely related to the learning of unknown classes, and simply emphasizing the learning of unknown classes may hurt overall classification. We can assume an extreme case: when the supervised loss $\mathcal{L}_{CE}^l$ is negligible, SSOC degenerates into a clustering algorithm on unlabeled data, without any information about ground-truth classes. When performing label matching using the Hungarian maximum weight algorithm, incorrect matches between predicted labels and true labels increase, resulting in suboptimal results for unknown classes and overall classification. Therefore, we need to find an optimal set of loss balance weights.


\section{Conclusion}
    In this work, we propose SSOC to address the open-world SSL problem. SSOC autonomously learns the categories in the open world using a cross-attention mechanism and leverages pairwise similarity loss to extract information from unlabeled data, discovering novel classes through instance prediction and relationships. We demonstrate the effectiveness of SSOC on three benchmark computer vision datasets, where it outperforms state-of-the-art baseline methods. Moreover, SSOC exhibits excellent robustness when facing challenges such as limited labeled data and many novel classes.

\section*{Acknowledgment}
    National Key RD Program of China (2022YFF0712100), NSFC (62006118, 62276131), Natural Science Foundation of Jiangsu Province of China under Grant (BK20200460), Jiangsu Shuangchuang (Mass Innovation and Entrepreneurship) Talent Program. Young Elite Scientists Sponsorship Program by CAST, the Fundamental Research Funds for the Central Universities (NO.NJ2022028, No.30922010317, No.30923011007).
    
\bibliographystyle{IEEEtran}
\bibliography{SSOC.bib}
\end{document}